\documentclass[letterpaper, 10 pt, conference]{ieeeconf}  

\IEEEoverridecommandlockouts                              

\usepackage{hyperref}
\usepackage{url}
\usepackage{amsmath}
\usepackage{amssymb}       
\usepackage{booktabs}
\usepackage{multirow}
\usepackage{graphicx}
\usepackage{subfig}
\title{AutoRSR: Automated Real-to-Sim-to-Real Data Generation for \\Scalable Robot Learning}

\author{
\textbf{Jiawei Zhang, Chang Liu, Litao Deng, Tao Zhang, Qin Jin}\\
Renmin University of China
\small
}

\begin{document}

\maketitle
\thispagestyle{empty}
\pagestyle{empty}

\begin{abstract}
Scaling embodied robot learning requires large and diverse manipulation datasets,
but collecting real-world demonstrations remains expensive in human effort and
robot time. 
Simulation provides a scalable alternative,
yet converting a real workspace into a robot-learning-ready environment still
requires substantial manual effort for object extraction, asset processing,
coordinate calibration, and trajectory specification. We present
\textbf{AutoRSR}, a highly automated real-to-sim-to-real framework that converts scanned real-world workspaces into reusable embodied simulation assets and scalable manipulation data. Given multi-view observations, robot states,
and open-vocabulary semantic queries, {AutoRSR} reconstructs the workspace,
extracts individual object assets, and automatically registers the reconstructed scene to the robot base
frame at metric scale. Starting from a small number of reference motions, AutoRSR then generates executable manipulation trajectories and scales them through trajectory interpolation and domain randomization.
After workspace scanning and semantic queries, the remaining pipeline requires no manual object annotation, scene-to-robot calibration, or robot teleoperation. Across three
real-world manipulation tasks, policies trained on 1K automatically generated
episodes per task achieve 73.3--86.7\% real-world success rates, comparable to $\pi_{0.5}$ trained from 150 human teleoperated demonstrations, while reducing the human time required for data generation by
an order of magnitude. These results demonstrate that real-world observations
can be converted into metrically aligned, executable, and scalable simulation
data with substantially reduced human effort. 

\end{abstract}

\section{Introduction}
\label{sec:introduction}

\begin{figure*}[!t]
\includegraphics[width=\textwidth]{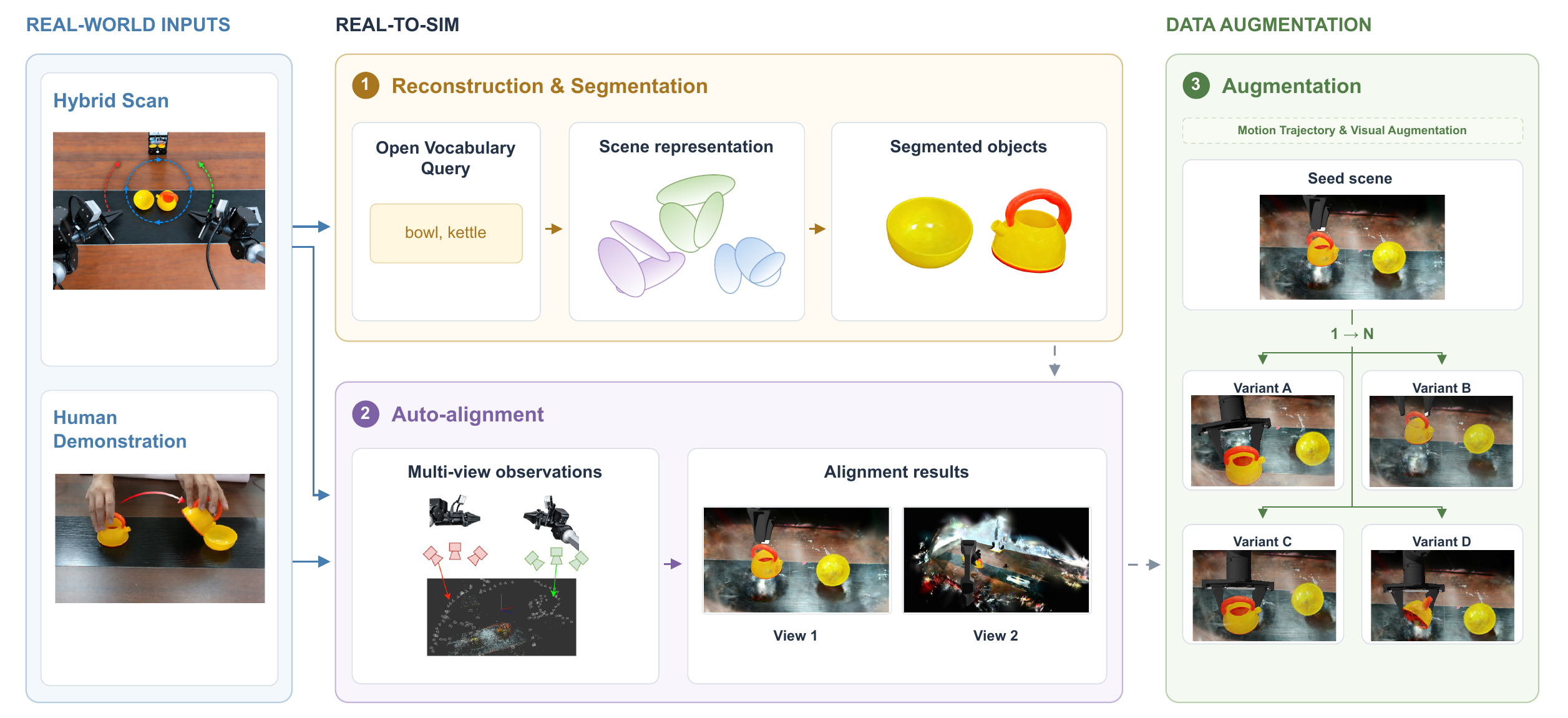}
\caption{Overview of \textbf{AutoRSR}. We first perform a hybrid scan of the workspace while recording robot states and wrist-camera poses, together with a reference manipulation motion. Given open-vocabulary semantic queries,
AutoRSR reconstructs the scene and extracts individual object assets. The reconstructed workspace is then automatically registered to the robot base frame, enabling reference trajectories to be replayed and augmented in simulation. Finally, trajectory interpolation and domain randomization generate scalable manipulation data for robot policy learning.
} \label{fig:overview}
\end{figure*}

A major bottleneck in embodied robot learning is the lack of scalable and diverse manipulation data. Real-world teleoperation provides high-fidelity demonstrations but requires substantial human effort and robot time, whereas simulation can generate large-scale data at much lower cost. The challenge, however, is not simply to create a visually realistic simulation.
For simulation to serve as a practical data engine, a real workspace must be converted into a \emph{robot-learning-ready embodied environment}: its objects should be individually reusable, its geometry should remain metrically consistent with the physical workspace, and its manipulation trajectories should be executable and scalable without repeatedly involving a human operator.


Existing approaches for constructing robot-learning simulations can be broadly categorized into generative and reconstruction-based methods. Generative
approaches~\cite{scenemaker,FlexWorld} offer flexible scene and asset synthesis, but the generated environments may differ from the geometry and appearance of a target workspace.
Reconstruction-based approaches~\cite{polaris,r2r2r,re3sim,splatsim,gase}
, in contrast, preserve the visual and spatial characteristics of real environments and are therefore attractive for sim-to-real robot learning. 
However, when reconstruction is used as a data generation engine, substantial human intervention remains between scene reconstruction and executable robot learning data. In particular, the remaining effort arises from two coupled challenges: preparing reusable simulation assets and making the reconstructed environment executable by a robot.


On \emph{reusable asset construction}:  
Obtaining simulation-ready objects from a real workspace typically involves separate object and scene scanning~\cite{polaris}, manual object segmentation~\cite{r2r2r}, or additional per-object processing for extraction and completion~\cite{gase}. 
Such operations introduce object-level human effort that grows with scene complexity and must be repeated
when the workspace changes. More importantly, these pipelines generally treat scene reconstruction and object preparation as separate stages rather than
providing a unified mechanism for turning a scanned workspace into reusable manipulation assets.

On \emph{executable trajectory generation}: A reconstructed scene alone is also insufficient for robot policy learning. The reconstruction is typically represented in its own coordinate system and must be registered to the robot before trajectories can be executed. This registration often requires explicit scene-to-robot calibration or manual alignment. Moreover, demonstrations used for learning are commonly collected through real-robot teleoperation, or trajectories must be manually specified and adapted to the reconstructed scene. Consequently, even when simulation removes the cost of physically executing thousands of demonstrations, human effort remains embedded in the process of producing the initial executable data. This limits the extent to which reconstruction-based simulation can serve as an automated and reusable data engine.

We address these limitations with \textbf{AutoRSR}, a highly automated real-to-sim-to-real framework that converts a scanned real workspace into scalable manipulation data. The key idea is to exploit information already
available during workspace scanning to eliminate subsequent manual alignment and annotation. We perform a hybrid scan of the workspace while recording wrist-camera poses and robot joint states. Given open-vocabulary semantic queries, these observations are used to reconstruct the workspace and extract
individual object assets without manually annotated masks. Because the camera is rigidly attached to the robot and its poses are synchronized with the robot
states, the reconstructed scene can be automatically registered to the robot
base frame at metric scale using the recorded geometric relationships. This produces a simulation environment in which reconstructed objects and robot motions are expressed in a physically consistent coordinate system.

Within this aligned environment, we use a human demonstration as motion seeds and automatically recover executable manipulation trajectories. Trajectory interpolation and domain randomization then produce diverse simulation episodes without requiring additional real-robot interaction. The result is a complete data-generation pipeline in which human effort is
concentrated on the initial workspace scan and semantic specification, while object extraction, coordinate registration, trajectory generation, and data
augmentation are automated.

We evaluate AutoRSR on three real-world manipulation tasks. For each task, we
generate 1K simulation episodes and train robot policies using only the automatically generated data. We compare them against policies trained from 150 human teleoperated real-world demonstrations. On the physical robot, AutoRSR-trained policies achieve 73.3--86.7\% success, with performance within
10 percentage points of the corresponding real-data baselines. We further measure the human time required by the complete data-generation process and observe an order-of-magnitude reduction compared with collecting an equivalent amount of real-world data. These results suggest that automatically reconstructed real workspaces can provide a practical intermediate representation for scalable robot learning.

In summary, our main contributions are: (i) We present \textbf{AutoRSR}, an automated real-to-sim-to-real pipeline that converts scanned real workspaces into reusable, metrically
aligned, and robot-executable simulation environments for manipulation data generation.
(ii) We introduce an automated processing for open-vocabulary object extraction, scene-to-robot registration, and executable trajectory
generation, enabling large-scale simulation data generation without manual object annotation, manual coordinate alignment, or robot teleoperation. (iii) We demonstrate the effectiveness of AutoRSR on three real-world manipulation tasks. Policies trained on 1K automatically generated episodes achieve real-robot performance comparable to policies trained on 150 human teleoperated  demonstrations, while substantially reducing human data-collection effort.

\section{Related Works}
\label{sec:related_work}
\subsection{Robot Data Collection Paradigms.}
The most prevalent source of robot manipulation data today is human teleoperation data\cite{aloha,aloha2,bridge,bridgedatav2,DROID,openx,RH20T}. However, collecting such data requires extensive human effort and time, making large-scale data acquisition prohibitively expensive. To obtain large volumes of robot manipulation data at a lower cost, researchers have explored various data generation paradigms.
Ego-centric\cite{Ego4D,egovlp} and humanoid-based approaches\cite{DeepMimic,h2O} consider human behaviors as a rich source of robot data. They enable robots to learn object interactions or joint motions through first-person hand manipulation data or human motion capture. UMI-based approaches\cite{umi,dexumi} enable fast and convenient collection of in-the-wild manipulation demonstrations by using a handheld gripper for human demonstrations and recovering the corresponding trajectories. Nevertheless, the embodiment gap between humans and robots remains a significant challenge.
In contrast, generating large-scale data through simulation augmentation offers the most cost-effective and convenient solution. MimicGen\cite{mimicgen,dexmimicen} generates diverse trajectories by transferring, perturbing, replaying, and composing a small number of human demonstrations. Other approaches, such as RoboGen\cite{robogen} and HRIBench\cite{HRIBench}, leverage large language models to plan behaviors and automatically generate simulated data. However, the sim-to-real gap introduced by simulation still limits the effectiveness of such data for training real-world robots.

\subsection{Real2Sim2Real Data Generation.}
To reduce the sim-to-real gap of simulation data, many approaches have begun to follow the \textit{Real2Sim2Real paradigm}, which reconstructs the real world from observations, imports the reconstructed environments into simulators, and then collects and augments large-scale data in simulation. After training, the learned policies are transferred back to the real world for inference\cite{polaris,r2r2r,re3sim,splatsim,gase}. With the rapid development of 3D Gaussian Splatting (3DGS)\cite{3dgs} and the increasing support for neural Gaussian representations in modern simulators, the visual sim-to-real gap has been significantly reduced. As a result, many approaches have started to build digital twins\cite{twinaligner} by combining neural Gaussian representations of objects for high visual fidelity with collision mesh representations for physical interactions within simulation environments.


\section{Method}
\label{sec:method}
We perform hybrid scene scanning using the robot-mounted wrist camera and third-party cameras while recording the robot states. We then use a head-mounted camera to capture a human manipulation demonstration video. Then we perform anchor-based semantic reconstruction. After separating the objects and the scene according to the user’s open-vocabulary semantic queries, we import them separately into the simulator. We then align the robot coordinate system with the reconstruction coordinate system, retarget the robot manipulation trajectory based on the demonstration video, and augment the aligned trajectories to generate a large-scale dataset. Users only need to provide scene-scanning data, a demonstration video, and open-vocabulary semantic queries for the target objects; the entire remaining pipeline is then executed fully automatically.

\subsection{Initialization} 
\label{subsec:Initialization}
To facilitate conversion of the scan data into the COLMAP format commonly used by 3DGS prior to reconstruction, we configured the two robotic arms to scan around the scene in a predefined sequence, with the left arm scanning first, followed by the right. Each arm’s return to the vicinity of its home position marked the completion of its data acquisition routine. We then established a mapping between image filenames and the corresponding robot poses to support subsequent alignment. We then process the combined set of scan images using COLMAP\cite{sfm,mvs} or PostShot\cite{postshot} to obtain a sparse point cloud and camera poses. The resulting camera poses are expressed in the reconstruction coordinate frame rather than the robot coordinate frame.

After scanning, we establish consistent semantic annotations across all input views. Given a selected reference frame, we use SAM2\cite{sam,sam2} to obtain object masks and propagate the corresponding object IDs throughout the video sequence, producing consistent semantic masks ${M_i}$ across views. We then transfer these 2D semantic labels to the initial sparse point cloud. Specifically, each point $\mathbf a$ is projected onto all visible views according to the estimated camera poses and depth-occlusion relationships. Semantic label is determined by majority voting over the masks observed from valid projections:
\begin{equation}
\ell_a =
\arg\max_{\ell \in V_\ell}
\sum_i v_{i,a}\,
\mathbb{I}\!\left[
M_i\!\left(\pi_i(\mathbf a)\right)=\ell
\right].
\end{equation}
where $\pi_i(\cdot)$ denotes projection onto the $i$-th view, $v_{i,a}$ indicates whether anchor $\mathbf a$ is visible in that view, and $V_\ell$ denotes the set of candidate semantic labels.

The labeled sparse points are subsequently voxelized to initialize the anchor representation. Each point $\mathbf a$ is quantized into an integer voxel coordinate $\mathbf q=\operatorname{round}(\mathbf a/h)$, where $h$ denotes the voxel size. Points assigned to the same voxel are merged into a single anchor located at the corresponding voxel-grid position. Since the initial point cloud remains sparse, particularly around thin structures and object boundaries, we further densify the representation using semantic-constrained anchor growing. At each iteration, we retrieve the KNN neighbors of each anchor and generate midpoint anchors only between semantically consistent anchor pairs. The newly generated anchors inherit the corresponding semantic label, are merged with the existing anchor set, and are subsequently re-voxelized to remove duplicates. Repeating this procedure progressively densifies sparsely represented regions while reducing anchor propagation across semantic boundaries, providing a cleaner and denser initialization for subsequent semantic Gaussian reconstruction.

\begin{figure}[!t]
    \centering
    \includegraphics[width=\columnwidth]{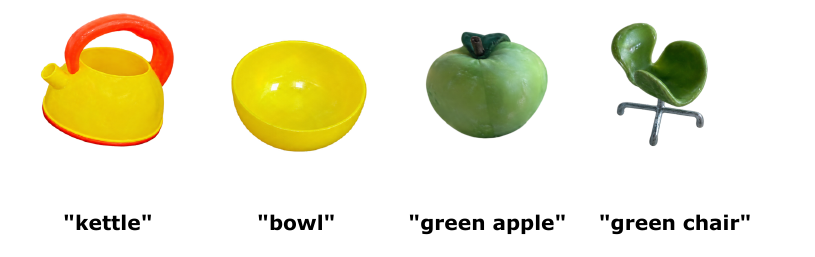}
    \caption{Object segmentation results on our \textit{Pour into the yellow bowl} task and LERF\cite{lerf} dataset.}
    \label{fig:objects}
\end{figure}

\subsection{Scene Reconstruction and Object Segmentation}
\subsubsection{Object-Aware Gaussian Reconstruction and Optimization}
\label{subsec:Reconstruction}
We adopt ObjectGS\cite{objectgs}, an object-aware anchor representation method following Scaffold-GS\cite{scaffoldgs}. Each anchor $a$ contains a spatial position $
  \mathbf{x}_a^W$, a learnable feature $\mathbf{f}_a$, an instance label $\ell_a$, a local scale, and $M$ learnable offsets. For the $i$-th camera, the anchor feature and viewing direction are jointly fed into opacity, color, and covariance MLPs to generate multiple neural Gaussians. Let $g=(a,m)$ denote the $m$-th Gaussian generated from anchor $a$, whose center is:
\begin{equation}
\boldsymbol\mu_g^W=\mathbf x_a^W+\mathbf s_a^{\mathrm{pos}}\odot\mathbf o_{a,m},\qquad g=(a,m),\label{eq:neural_gaussian_center}
\end{equation}
,where $\mathbf{o}_{a,m}$ is the normalized local offset and $\mathbf{s}_a^{\mathrm{pos}}$ denotes the positional scale of anchor $a$. The Gaussian attributes, including color, opacity, covariance scale, and rotation, are predicted by their respective MLPs. 

  At each training iteration, we sample a viewpoint and perform differentiable Gaussian rasterization to obtain RGB images, alpha maps, and semantic scores. The
  reconstruction objective is
  \begin{multline}
  \mathcal{L}_{\mathrm{rec}}
  =(1-\lambda_{\mathrm{ssim}})\mathcal{L}_1
  +\lambda_{\mathrm{ssim}}(1-\operatorname{SSIM})\\
  +\lambda_{\mathrm{obj}}\mathcal{L}_{\mathrm{CE}}
  +\mathcal{L}_{\mathrm{base}},
  \label{eq:reconstruction_loss}
  \end{multline}
  where $\mathcal{L}_{\mathrm{CE}}$ supervises the rendered semantic scores using the 2D object-ID masks, ignoring background label $0$. $\mathcal{L}_{\mathrm{base}}$
  collects the auxiliary penalties enabled in the configuration, including scale, depth, normal, distortion, opacity, and background-semantic regularization.
  
  During densification of Gaussian training, accumulated screen-space gradients of child Gaussians identify regions requiring additional representation capacity. New anchors are generated in
  multi-scale voxel grids and inherit the label of the parent with the strongest gradient contribution. All child Gaussians spawned from an anchor share the one-hot semantic vector associated with $\ell_a$, coupling object
  identity with local geometric generation and enabling direct instance selection.

\subsubsection{Object Extraction and Asset Generation}
\label{subsec:Extraction}
Given a user query, we use GroundingDINO\cite{groundingdino} and SAM to obtain a target mask on a rendered view, then associate it with reconstructed instance IDs
  through overlap scoring against the rendered semantic-ID map. We extract the corresponding anchors according to the selected IDs. For object removal, we expand the removal region through spatial neighborhoods to suppress residual related Gaussians, yielding both \textbf{object-only} and
  \textbf{scene-without-object} representations. Since anchor-based Gaussian attributes are view-dependent, we evaluate their MLPs from multiple training cameras and
  aggregate the decoded attributes into standard explicit 3DGS assets. Object export uses opacity-based cross-view consensus and median aggregation, while scene export
  averages colors and opacities over accepted views to preserve coverage. When necessary, the explicit Gaussians are further distilled against multi-view renders from the
  original anchor model to reduce conversion artifacts.

  To obtain simulation geometry, we use SuGaR\cite{sugar} to reconstruct meshes from the explicit Gaussians. For isolated objects, anisotropic Gaussian support removes
  unsupported mesh faces, followed by voxelization and surface reconstruction to generate watertight collision meshes. When articulated assets are required, the resulting
  meshes can be further processed by Articulate-Anymesh\cite{Articulate_AnyMesh}.

\begin{figure}[!t]
    \centering
    \includegraphics[width=\columnwidth]{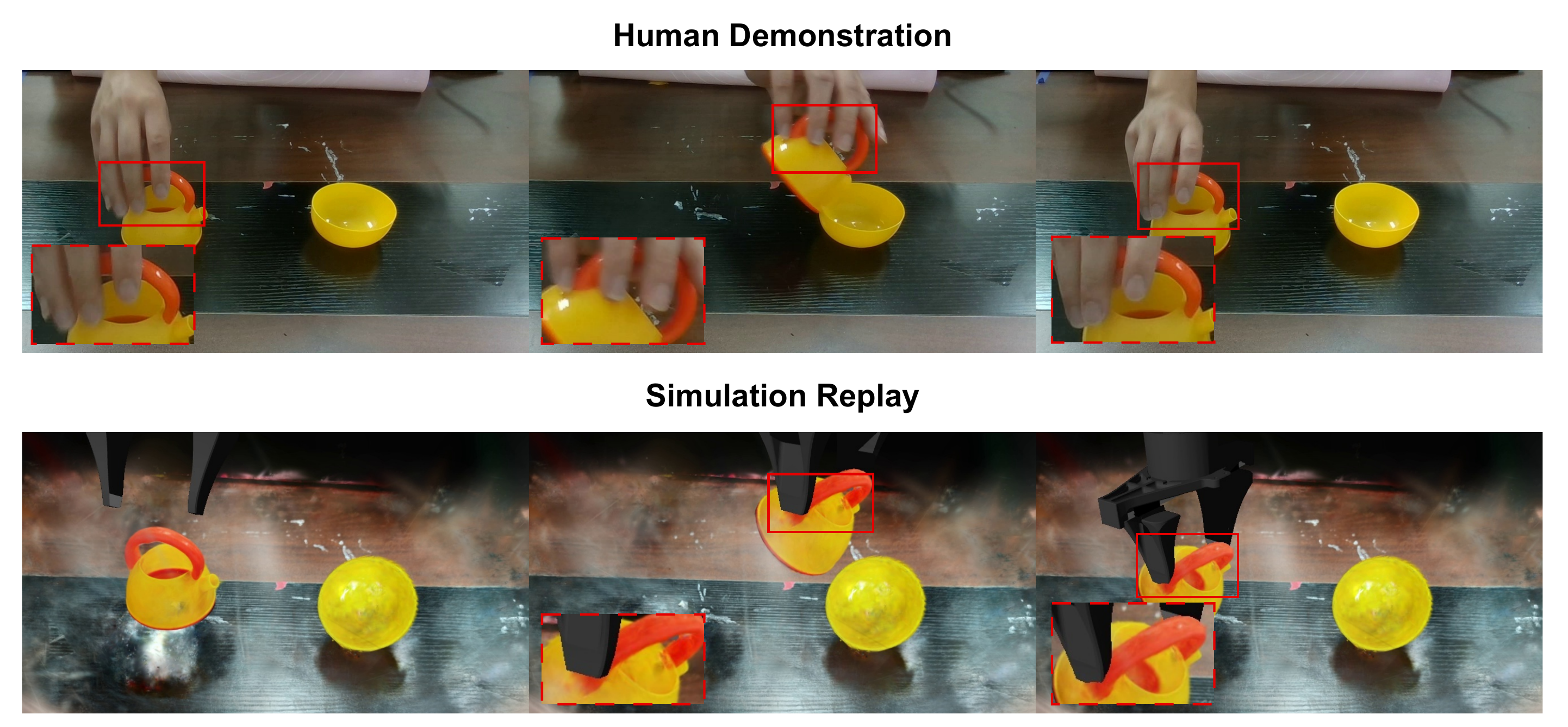}
    \caption{Example of demonstration alignment for the \textit{Pour into the yellow bowl} task.}
    \label{fig:replay}
\end{figure}

\subsection{Demonstration Alignment and Motion Retargeting}
  \label{subsec:Alignment}

  \subsubsection{Metric Coordinate Alignment}
  We align the reconstructed scene with the robot using calibrated wrist-camera poses collected during scanning. For each image, the metric camera pose is computed from
  the synchronized robot configuration $\mathbf{q}_i$, robot-base extrinsics, and hand--eye calibration:
  \begin{equation}
      \mathbf{T}_{RC_i}
      =
      \mathbf{T}_{RB}
      \mathbf{T}_{BE}(\mathbf{q}_i)
      \mathbf{T}_{EC},
  \end{equation}
  where $\mathbf{T}_{AB}$ maps coordinates from frame $B$ to frame $A$, and $R$, $B$, $E$, and $C$ denote the robot rig, robot base, wrist mounting frame, and camera
  frame, respectively. Corresponding reconstructed and robot-derived camera poses determine a similarity transform through RANSAC followed by robust refinement using
  camera positions and orientations. Object poses are then mapped from reconstruction coordinates $W$ into metric robot coordinates:
  \begin{equation}
          \mathbf{p}^{R}
          = s\mathbf{R}_{RW}\mathbf{p}^{W}+\mathbf{t}_{RW},
          \mathbf{R}_{RO}
          = \mathbf{R}_{RW}\mathbf{R}_{WO}.
  \end{equation}
  The same scale is applied to Gaussian assets and collision meshes. This transformation is performed after reconstruction, leaving the original COLMAP model unchanged.
  The demonstration is interpreted using the calibrated head-camera pose, rather than assigning it to an arbitrary scanning image.

  \subsubsection{Object Tracking and Grasp Retargeting}
  Using the open-vocabulary semantic queries already provided by the user, we employ GroundingDINO\cite{groundingdino} and SAM2\cite{sam2} to locate and segment the target object in the first frame of the human demonstration video, then propagate its mask throughout the video. Since all required inputs have already been prepared in the preceding stages, this process is fully automated.

We then recover the object’s motion in 3D from the 2D demonstration video through differentiable Gaussian rendering. Gaussian appearance and local geometry remain fixed while the object pose is optimized using a foreground–background balanced silhouette loss, Dice loss, and robust foreground photometric error. Temporal regularization and a bounded image-centroid prior discourage abrupt changes in tracking. Initialization is constrained to a neighborhood of the reconstructed object pose, and subsequent frames use the preceding estimate as their initial guess.

Next, we extract object geometry from the collision mesh and generate grasp candidates with different approach directions, gripper-closing directions, and positions. Candidate generation accounts for the gripper’s opening range and the object geometry. We use image-space hand cues provided by HaMeR~\cite{HaMeR} to rank grasp candidates according to how closely their projected grasp locations match the thumb–index midpoint in the demonstration. These cues guide grasp selection without treating monocular hand-depth estimates as metric ground truth.

We plan both the gripper opening at contact and its opening during approach. The desired end-effector motion is determined by the tracked object motion and the selected grasp pose in the object’s local frame. Inverse kinematics (IK) with joint-limit constraints is solved using sequential warm starts along the trajectory, and candidate acceptance checks both position and orientation residuals over the complete sequence.

Robot forward kinematics, visual meshes, and collision geometry use the same URDF link transforms. During rigid-object transport with the gripper fully closed, the object pose is bound to the achieved end-effector pose while preserving the local grasp transform. Gaussian and collision geometry therefore move together despite finite IK residuals. At release, the achieved placement pose is retained, rather than snapping the object back to the nominal trajectory or allowing it to follow the retreating gripper. For constrained articulated or pushing motions, the prescribed object kinematics are retained instead, and contact is checked against the achieved robot motion.

  \subsection{Motion Trajectory and Visual Augmentation}
  \label{subsec:Augmentation}

  \subsubsection{Motion Trajectory Augmentation}
  We construct a reusable bank of feasible motion templates before expanding appearance diversity.  We independently randomize the tabletop placements of all objects specified for segmentation by the open-vocabulary semantic queries. We also modify their geometric properties. Planar translations are sampled from a bounded distribution, while yaw rotations are applied about each object’s initial position rather than the world origin. Initial heights remain unchanged, and the system preserves the validated grasp relationships and object scales.

  Independent placement is coupled with task-relative trajectory adaptation. The manipulated object retains its sampled initial pose, while its terminal pose is adjusted
  to preserve the demonstrated relationship to the independently placed receiver. A smooth time-dependent correction blends positions directly in world coordinates and
  interpolates the relative orientation. An explicitly specified interaction interval can enforce the relationship throughout that interval instead of only at the
  endpoint. This preserves the demonstrated motion structure while avoiding unintended orbital motion around the coordinate origin.

  Each candidate is expanded into a synchronized approach, gripper-closing, and task-motion sequence. For placement tasks, a geometric support test determines whether a
  short settling motion, gripper opening, and retreat can be appended. Support is assessed using object-relative surface probes, so a common planar transformation does
  not change the verdict. An object without confirmed terminal support remains held and required-placement candidates are rejected if support cannot be established. Settling
  is a bounded contact adjustment, not randomized vertical placement.

  Motion candidates are solved in batches, with sequential warm starts within each trajectory. We screen grasp support, joint limits, and pose residuals, and recheck
  terminal support after binding rigid objects to achieved forward kinematics. Replay additionally checks finger contact during fully closed intervals and excessive
  robot--object mesh penetration. Collision meshes follow their corresponding Gaussian assets but contribute no object appearance. These checks provide geometric
  feasibility screening, rather than a guarantee of force closure or dynamic task success.

  \begin{figure*}[!t]
    \centering
    \includegraphics[width=\linewidth]{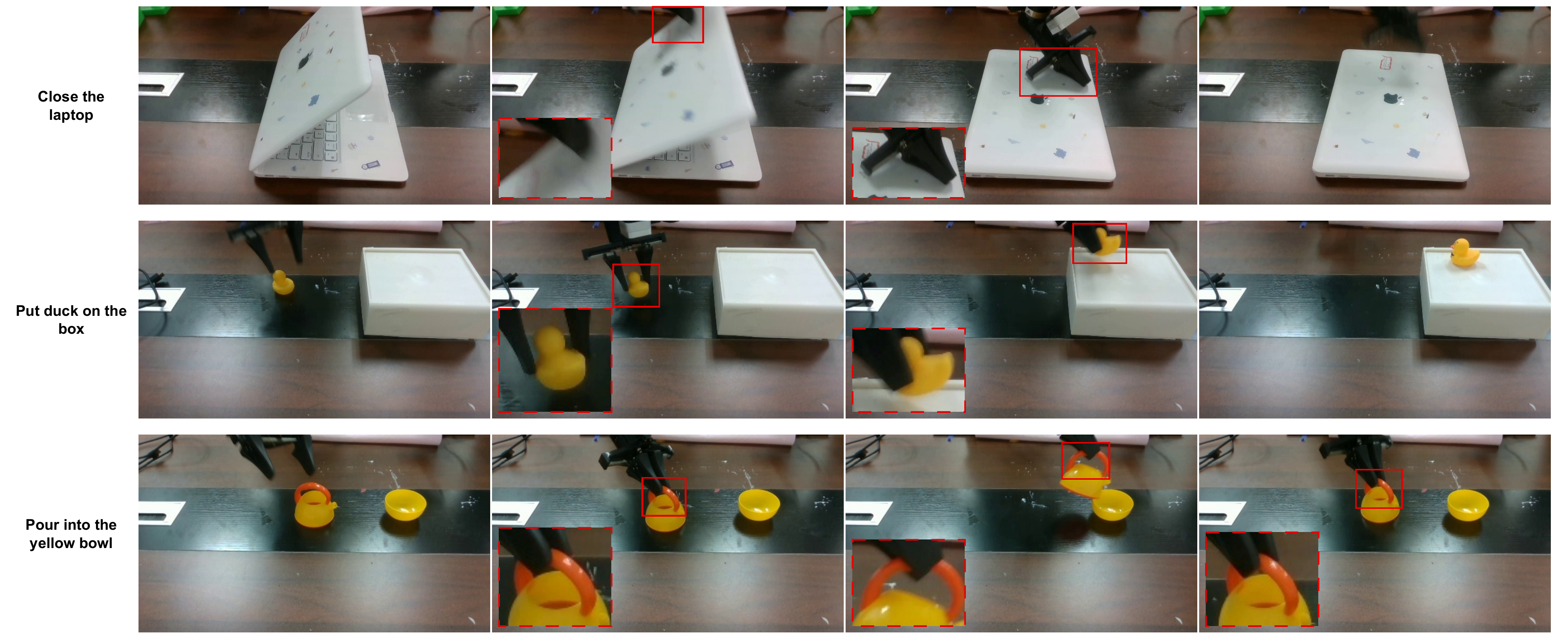}
    \caption{Real-robot inference results for $\pi_{0.5}$\cite{pi0.5} fine-tuned on 1,000 AutoRSR-augmented episodes. The solid red box highlights the target manipulation region, and the dashed red box at the bottom‑left provides a magnified view of this region for better visualization}
    \label{fig:manipulation}
\end{figure*}

  \subsubsection{Visual Augmentation}
 We combine the large number of augmented motion templates with multiple visual configurations. While keeping the geometry, motion, actions, and timing unchanged, we randomize the textures of both objects and scenes, and apply coherent adjustments to the Gaussians’ exposure and white balance. Lighting variation is applied to the robot mesh renderer.

During rendering, we continuously reuse Genesis\cite{genesisworld,Genesis} runtime resources and cached Gaussian assets, and batch-process both Gaussian data and camera requests for compatible episodes. Robot meshes and Gaussian foreground/background layers are rendered from consistent camera poses and composited using depth. The head camera remains fixed in the rig frame, while wrist-camera poses are recomputed from the corresponding robot joint configurations and hand–eye transformations.

All outputs share a single episode timeline: camera observations, proprioceptive states, robot actions, and gripper openings all use aligned frame indices and timestamps. Each episode contains the observations and actions required for training. The observation views retain the input robot wrist-camera view and head-camera view.

\section{Experiments}
\label{sec:experiments}

\subsection{Setups}
We perform hybrid scene scanning using the built-in camera of a handheld smartphone and two wrist-mounted D405 cameras. In addition, a D435 camera is mounted at a fixed position in the scene to provide a top-down view and record human demonstration videos. The coordinate frames of both D405 cameras and the D435 camera are registered to the robot arm’s coordinate frame. The robot platform used in our experiments is a PiperX\cite{piperx} robotic arm. All experiments were conducted on a single NVIDIA A800 GPU. For real-robot manipulation experiments, we designed three tasks and generated 1,000 augmented trajectories for each task using our system. For each of the three tasks, we also collect 150 teleoperated demonstrations. We fine-tune the $\pi_{0.5}$\cite{pi0.5} policy using LoRA\cite{lora} for 30,000 steps with a batch size of 32 separately on the teleoperated and augmented datasets and compare the resulting manipulation success rates. During fine-tuning, we update all parameters of the vision encoder and visual projection layer in $\pi_{0.5}$, and apply LoRA to the attention and MLP layers of both the language model and the action expert. The language embeddings and normalization layers remain frozen. We use a learning rate of $2.5 \times 10^{-5}$ and an action horizon of 10.

\begin{figure*}[!t]
    \centering
    \includegraphics[width=\linewidth]{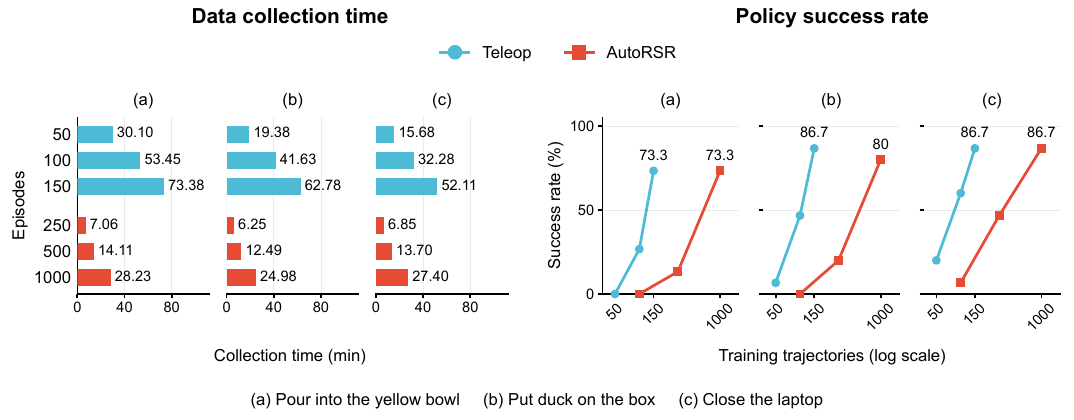}
    \caption{Comparison of data collection time and physical policy success rates between teleoperation and AutoRSR.}
    \label{fig:data_collection_time}
\end{figure*}

\begin{table}[t]
\centering\small
\caption{Comparison of Physical Policy Success Rates Across Training Sources(\%).}
\label{tab:physical_policy_success}
\renewcommand{\arraystretch}{1.05}
\setlength{\tabcolsep}{8pt}

\resizebox{\columnwidth}{!}{%
\begin{tabular}{l|ccc|ccc}
\toprule
\textbf{Task}
& \multicolumn{3}{c|}{\textbf{Teleop Trajectories}}
& \multicolumn{3}{c}{\textbf{AutoRSR Trajectories}} \\
\cmidrule(lr){2-4}
\cmidrule(lr){5-7}
& 50 & 100 & 150
& 100 & 300 & 1000 \\
\midrule

\textbf{Pour into the yellow bowl}
& 0 & 26.7 & 73.3
& 0 & 13.3 & 73.3 \\
\midrule

\textbf{Put duck on the box}
& 6.7 & 46.7 & 86.7
& 0 & 20 & 80 \\
\midrule

\textbf{Close the laptop}
& 20 & 60 & 86.7
& 6.7 & 46.7 & 86.7 \\

\bottomrule
\end{tabular}%
}
\end{table}

\begin{table}[!t]
    \centering
    \small
    \caption{Training-data mixture comparison(\%).}
    \label{tab:cotrain_comparison}
        \begin{tabular}{ccc}
            \toprule
            \textbf{Real-only (150)}
            & \textbf{Sim-only (1k)}
            & \textbf{Co-Train (150+1k)} \\
            \midrule
            73.3
            & 73.3
            & \textbf{86.7} \\
            \bottomrule
        \end{tabular}%
\end{table}

\subsection{Manipulation}
We design three tasks: (1)\textbf{ Pour into the yellow bowl}. Grasp the plastic jug, move it above the yellow bowl, and perform a pouring motion. (2) \textbf{Put the duck on the box.} Grasp the toy duck and place it on the white box. (3) \textbf{Close the laptop.} Close the lid of the open laptop model.

\paragraph{Raw Evaluation Results}
For each task, we collect 150 teleoperated demonstrations and then generate 1,000 augmented episodes using AutoRSR.  We fine-tune the $\pi_{0.5}$\cite{pi0.5} policy using LoRA on teleoperated datasets comprising the first 50, 100, or 150 episodes and simulation-augmented datasets comprising the first 100, 300, or 1,000 episodes. Each trained policy is then deployed on the real robot for evaluation. During deployment, we randomize object placements in the scene and introduce slight variations in lighting and geometric conditions. We conduct 15 trials per task and calculate the success rate. We also conducted a co-training experiment on the \textit{Pour into the yellow bowl} task using a mixture of 150 teleoperated demonstrations and 1,000 simulated demonstrations. The results in Tab.\ref{tab:physical_policy_success} show that the success rate achieved with 1,000 simulated samples is within 10 percentage points of that achieved with 150 real-robot samples. Tab.\ref{tab:cotrain_comparison} also indicate that co-training can improve the success rate.

\paragraph{Ablation}
We conducted separate ablation studies on motion trajectory augmentation and visual augmentation. For motion trajectory augmentation, we evaluated a setting in which it was completely disabled, replaying only the original object motion trajectories. For visual augmentation, we reduced its strength rather than disabling it entirely, as it was difficult to generate a sufficient amount of valid data using motion trajectory augmentation alone. The results are reported in the Tab.\ref{tab:trajectory_interpolation} and Tab.\ref{tab:visual_augmentation}. When motion trajectory augmentation is disabled, the model fails to generalize because it is trained on too few trajectories. In contrast, reducing visual augmentation degrades the model’s generalization ability but does not lead to a catastrophic loss of this capability.

\begin{table}[!t]
    \centering
    \small
    \caption{Motion trajectory augmentation ablation(\%).}
    \label{tab:trajectory_interpolation}
        \begin{tabular}{cc}
            \toprule
            \textbf{w/o Traj. Augmentation}
            & \textbf{w/ Traj. Augmentation} \\
            \midrule
            13.3
            & \textbf{73.3} \\
            \bottomrule
        \end{tabular}%
\end{table}

\begin{table}[!t]
    \centering
    \small
    \caption{Visual augmentation ablation(\%).}
    \label{tab:visual_augmentation}
        \begin{tabular}{cc}
            \toprule
            \textbf{less Vis. Augmentation}
            & \textbf{more Vis. Augmentation} \\
            \midrule
            60
            & \textbf{73.3} \\
            \bottomrule
        \end{tabular}%
\end{table}

\subsection{Alignment accuracy}
To evaluate the accuracy of our trajectory alignment, we recreated the scene layouts for the three tasks and placed a calibration board at ten randomly selected positions on the physical tabletop. For each placement, we computed the board’s corner coordinates in both our aligned coordinate frame and the coordinate frame obtained from the original COLMAP reconstruction via a Sim(3) transformation.
 We then measured the 3D positional errors relative to their ground-truth coordinates which are calculated in real workspace. As shown in the Tab.\ref{tab:alignment_errors}, our alignment achieves substantially lower mean error and root mean square error (RMSE) than the original reconstruction, with the estimated positions closely matching the ground truth.

\begin{table}[!t]
    \centering
    \caption{Alignment errors across different scenes (cm). (R) represents reconstruction coordinate frame and (A) represents our aligned coordinate frame.}
    \label{tab:alignment_errors}
    \small
    \renewcommand{\arraystretch}{1.2}
    \resizebox{\columnwidth}{!}{ \begin{tabular}{lcccc}
        \toprule
        task  & Mean Error(R) & Mean Error(A) & RMSE(R) & RMSE(A) \\
        \midrule
        grasp\_duck & 8.62 & \textbf{0.61} & 9.07 & \textbf{0.62} \\
        pour       & 4.58 & \textbf{0.91} & 5.02 & \textbf{0.91} \\
        computer   & 3.81 & \textbf{0.77} & 3.95 & \textbf{0.78} \\
        \bottomrule
    \end{tabular}}
\end{table}

\begin{table}[!t]
\centering
\caption{Stage-by-stage comparison of human involvement across the entire pipeline.\textbf{ H} indicates that human involvement is required, while \textbf{A} indicates that the stage is completed automatically by the system. Since GASE does not perform trajectory reproduction or augmentation, its corresponding entries in the demonstration and augmentation sections are left blank.}
\label{tab:human_time}
\begin{tabular}{lcc|c}
\toprule
Task & r2r2r\cite{r2r2r} &GASE\cite{gase} & Ours\\
\midrule
Scanning              &  H & H    &H \\
Demonstration         &   H  &-  &H\\
Query                 & H  &  H & H\\
Reconstruction        & A  &A   & A\\
Segmentation          &H & H  &A \\
Assert into Simulation     & H &H &A \\
Alignment             &H  &H   &A \\
Augmentation         &   A  &-  &A\\
\bottomrule
\end{tabular}
\end{table}

\begin{table}[!t]
\centering
\caption{Upfront processing time until generation. We train the reconstruction model for 7,000 iterations and directly extract meshes using SuGaR\cite{sugar}.}
\label{tab:processing_time}
\begin{tabular}{lc}
\toprule
Task & Time to complete\\
\midrule
Scanning                    & 1 min\\
Demonstration       &10 s\\
Mask processing              & 2 min\\
Reconstruction                & 10 min\\
Segmentation               &$<$10 s \\
Meshification                & 2 min\\
Alignment            &   10 min\\ 
\bottomrule
\end{tabular}
\end{table}

\begin{table}[!t]
    \centering\small
    \caption{
        Time required for collecting 150 demonstrations through teleoperation
        with one human operator and for generating 1,000 synthetic demonstrations
        using AutoRSR with single GPU.
    }
    \label{tab:data_collection_time}
    
    \renewcommand{\arraystretch}{1.1}
    \setlength{\tabcolsep}{10pt}
    \resizebox{\columnwidth}{!}{
     \begin{tabular}{@{}lcccc@{}}
        \toprule
        \textbf{Task}
        & \textbf{Teleop}
        & \textbf{AutoRSR(Aug)}\\
        \midrule

        Pour into the yellow bowl
        & 73.38 min
        & 28.23 min \\

        Put duck on the box
        & 62.78 min
        & 24.98 min           \\

        Close the laptop
        & 52.11 min
        & 27.40 min \\
        \bottomrule
    \end{tabular}
    }
\end{table}

\subsection{Time cost and human involvment comparison}
To demonstrate the efficiency of our system, we conduct time cost  and human involvment comparison. For each of the three manipulation tasks, we first compare the time required to collect 150 teleoperated demonstrations with that required to generate 1,000 augmented episodes using AutoRSR. As shown in the Tab.\ref{tab:data_collection_time}, AutoRSR generally requires no more than half the time needed to collect equivalent data through teleoperation. Moreover, teleoperated data collection requires continuous human involvement, whereas AutoRSR augmentation requires almost no human intervention.

In addition, we compare AutoRSR with other systems in terms of the procedures that require human involvement throughout the pipeline. As shown in Tab.\ref{tab:human_time} that our AutoRSR requires users only to provide scene scans, demonstration data, and open-vocabulary semantic queries. All remaining stages can be completed fully automatically, whereas other systems require substantial human intervention. In contrast, AutoRSR automates most of the pipeline. Finally, Tab.\ref{tab:processing_time} presents a breakdown of AutoRSR’s end-to-end processing time for the task \textit{Pour into the yellow bowl}.


\section{Conclusion}
\label{sec:conclusion}
In this paper, we present AutoRSR, a highly automated simulation data reconstruction and augmentation framework. The proposed system efficiently reconstructs real-world scenes, extracts object assets through semantic segmentation, aligns manipulation trajectories, and generates large-scale augmented data, enabling cost-effective acquisition of high-quality simulation data for policy learning. Real-robot experiments validate the effectiveness of our approach and demonstrate a small real-to-sim gap.






\bibliographystyle{IEEEtran} 
\bibliography{IEEEfull,IEEEexample}

@misc{aloha2,
      title={ALOHA 2: An Enhanced Low-Cost Hardware for Bimanual Teleoperation}, 
      author={Jorge Aldaco and Travis Armstrong and et.al},
      year={2024},
      eprint={2405.02292},
      archivePrefix={arXiv},
      primaryClass={cs.RO},
      url={https://arxiv.org/abs/2405.02292}, 
}

@misc{aloha,
      title={Learning Fine-Grained Bimanual Manipulation with Low-Cost Hardware}, 
      author={Tony Z. Zhao and Vikash Kumar and Sergey Levine and Chelsea Finn},
      year={2023},
      eprint={2304.13705},
      archivePrefix={arXiv},
      primaryClass={cs.RO},
      url={https://arxiv.org/abs/2304.13705}, 
}

@article{scenemaker,
        title={SceneMaker: Open-set 3D Scene Generation with Decoupled De-occlusion and Pose Estimation Model},
        author={Shi, Yukai and Li, Weiyu and et.al},
        journal={arXiv preprint arXiv:2512.10957},
        year={2025}
      }

@misc{FlexWorld,
      title={FlexWorld: Progressively Expanding 3D Scenes for Flexiable-View Synthesis}, 
      author={Luxi Chen and Zihan Zhou and et.al},
      year={2025},
      eprint={2503.13265},
      archivePrefix={arXiv},
      primaryClass={cs.CV},
      url={https://arxiv.org/abs/2503.13265}, 
}

@misc{polaris,
      title={PolaRiS: Scalable Real-to-Sim Evaluations for Generalist Robot Policies}, 
      author={Arhan Jain and Mingtong Zhang and et.al},
      year={2025},
      eprint={2512.16881},
      archivePrefix={arXiv},
      primaryClass={cs.RO},
      url={https://arxiv.org/abs/2512.16881}, 
}

@misc{r2r2r,
        title={Real2Render2Real: Scaling Robot Data Without Dynamics Simulation or Robot Hardware},
        author={Justin Yu and Letian Fu and et.al},
        year={2025},
        eprint={2505.09601},
        archivePrefix={arXiv},
        primaryClass={cs.RO},
        url={https://arxiv.org/abs/2505.09601},
  }

@inproceedings{re3sim,
  title={RE$^3$SIM: Generating High-Fidelity Simulation Data via 3D-Photorealistic Real-to-Sim for Robotic Manipulation},
  author={Han, Xiaoshen and Yu, Junqiu and et.al},
  booktitle={ICRA},
  year={2026}
}

@misc{gase,
  title={GASE: Gaussian Splatting-Based Automated System for Reconstructing Embodied-Simulation Environments},
  author={Jiawei Zhang and et.al},
  year={2026},
  eprint={2606.17520},
  archivePrefix={arXiv},
  primaryClass={cs.RO},
  url={https://arxiv.org/abs/2606.17520},
}

@misc{splatsim,
            title={SplatSim: Zero-Shot Sim2Real Transfer of RGB Manipulation Policies Using Gaussian Splatting}, 
            author={Mohammad Nomaan Qureshi and Sparsh Garg and et.al},
            year={2024},
            eprint={2409.10161},
            archivePrefix={arXiv},
            primaryClass={cs.RO},
            url={https://arxiv.org/abs/2409.10161}, 
      }

@Article{3dgs,
      author       = {Kerbl, Bernhard and Kopanas, Georgios and Leimk{\"u}hler, Thomas and Drettakis, George},
      title        = {3D Gaussian Splatting for Real-Time Radiance Field Rendering},
      journal      = {ACM Transactions on Graphics},
      number       = {4},
      volume       = {42},
      month        = {July},
      year         = {2023},
      url          = {https://repo-sam.inria.fr/fungraph/3d-gaussian-splatting/}
}

@article{sam,
  title={Segment Anything},
  author={Kirillov, Alexander and Mintun, Eric and et.al},
  journal={arXiv:2304.02643},
  year={2023}
}

@article{sam2,
  title={SAM 2: Segment Anything in Images and Videos},
  author={Ravi, Nikhila and Gabeur, Valentin and et.al},
  journal={arXiv preprint arXiv:2408.00714},
  url={https://arxiv.org/abs/2408.00714},
  year={2024}
}

@inproceedings{scaffoldgs,
  title={Scaffold-gs: Structured 3d gaussians for view-adaptive rendering},
  author={Lu, Tao and Yu, Mulin and et.al},
  booktitle={CVPR},
  pages={20654--20664},
  year={2024}
}

@inproceedings{objectgs,
  title={ObjectGS: Object-aware Scene Reconstruction and Scene Understanding via Gaussian Splatting},
  author={Zhu, Ruijie and Yu, Mulin and et.al},
  booktitle={ICCV, 2025},
  year={2025}
}

@misc{openx,
title={Open {X-E}mbodiment: Robotic Learning Datasets and {RT-X} Models},
author = {Open X-Embodiment Collaboration and et.al},
howpublished  = {\url{https://arxiv.org/abs/2310.08864}},
year = {2023},
}

@misc{bridge,
      title={Bridge Data: Boosting Generalization of Robotic Skills with Cross-Domain Datasets}, 
      author={Frederik Ebert and Yanlai Yang and et.al},
      year={2021},
      eprint={2109.13396},
      archivePrefix={arXiv},
      primaryClass={cs.RO},
      url={https://arxiv.org/abs/2109.13396}, 
}

@inproceedings{bridgedatav2,
    title={BridgeData V2: A Dataset for Robot Learning at Scale},
    author={Walke, Homer and Black, Kevin and et.al},
    booktitle={CoRL},
    year={2023}
}

@inproceedings{
  RH20T,
  title        = {RH20T: A Comprehensive Robotic Dataset for Learning Diverse Skills in One-Shot},
  author       = {Fang, Hao-Shu and Fang, Hongjie and et.al},
  booktitle    = {ICRA},
  pages        = {653--660},
  year         = {2024},
  organization = {IEEE}
}

@misc{DROID,
      title={DROID: A Large-Scale In-The-Wild Robot Manipulation Dataset}, 
      author={Alexander Khazatsky and Karl Pertsch and et.al},
      year={2025},
      eprint={2403.12945},
      archivePrefix={arXiv},
      primaryClass={cs.RO},
      url={https://arxiv.org/abs/2403.12945}, 
}

@misc{Ego4D,
      title={Ego4D: Around the World in 3,000 Hours of Egocentric Video}, 
      author={Kristen Grauman and Andrew Westbury and et.al},
      year={2022},
      eprint={2110.07058},
      archivePrefix={arXiv},
      primaryClass={cs.CV},
      url={https://arxiv.org/abs/2110.07058}, 
}

@inproceedings{egovlp,
 author = {Lin, Kevin Qinghong and Wang, Jinpeng and et.al},
 booktitle = {Advances in Neural Information Processing Systems},
 pages = {7575--7586},
 publisher = {Curran Associates, Inc.},
 title = {Egocentric Video-Language Pretraining},
 url = {https://proceedings.neurips.cc/paper_files/paper/2022/file/31fb284a0aaaad837d2930a610cd5e50-Paper-Conference.pdf},
 volume = {35},
 year = {2022}
}

@article{DeepMimic,
   title={DeepMimic: example-guided deep reinforcement learning of physics-based character skills},
   volume={37},
   ISSN={1557-7368},
   url={http://dx.doi.org/10.1145/3197517.3201311},
   DOI={10.1145/3197517.3201311},
   number={4},
   journal={ACM Transactions on Graphics},
   publisher={Association for Computing Machinery (ACM)},
   author={Peng, Xue Bin and Abbeel, Pieter and et.al},
   year={2018},
   month=July, pages={1–14} }

@misc{h2O,
      title={Learning Human-to-Humanoid Real-Time Whole-Body Teleoperation}, 
      author={Tairan He and Zhengyi Luo and et.al},
      year={2024},
      eprint={2403.04436},
      archivePrefix={arXiv},
      primaryClass={cs.RO},
      url={https://arxiv.org/abs/2403.04436}, 
}

@misc{robogen,
      title={RoboGen: Towards Unleashing Infinite Data for Automated Robot Learning via Generative Simulation}, 
      author={Yufei Wang and Zhou Xian and et.al},
      year={2023},
      eprint={2311.01455},
      archivePrefix={arXiv},
      primaryClass={cs.RO}
}

@inproceedings{mimicgen,
    title={MimicGen: A Data Generation System for Scalable Robot Learning using Human Demonstrations},
    author={Mandlekar, Ajay and Nasiriany, Soroush and et.al},
    booktitle={CoRL},
    year={2023}
}

@inproceedings{dexmimicen,
      title     = {DexMimicGen: Automated Data Generation for Bimanual Dexterous Manipulation via Imitation Learning},
      author    = {Jiang, Zhenyu and Xie, Yuqi and et.al},
      booktitle = {ICRA},
      year      = {2025}
}

@misc{HRIBench,
      title={HRIBench: Benchmarking Interaction-Centric Human-Robot Collaboration}, 
      author={Chang Liu and Jiawei Zhang and et.al},
      year={2026},
      eprint={2607.13056},
      archivePrefix={arXiv},
      primaryClass={cs.RO},
      url={https://arxiv.org/abs/2607.13056}, 
}

@article{
   genesisworld,
   author = {Genesis AI Team},
   title = {The Role of Simulation in Scalable Robotics, Genesis World 1.0, and the Path Forward},
   journal = {Genesis AI Blog},
   month = {May},
   year = {2026},
}

@misc{
  Genesis,
  author = {Genesis Authors},
  title = {Genesis: A Generative and Universal Physics Engine for Robotics and Beyond},
  month = {December},
  year = {2024},
  url = {https://github.com/Genesis-Embodied-AI/genesis-world}
}

@article{twinaligner,
  author    = {Hongwei Fan and Hang Dai and et.al},
  title     = {TwinAligner: Visual-Dynamic Alignment Empowers Physics-aware Real2Sim2Real for Robotic Manipulation},
  year={2025},
  eprint={2512.19390},
  archivePrefix={arXiv},
  primaryClass={cs.RO},
  url={https://arxiv.org/abs/2512.19390},
}

@article{sugar,
  title={SuGaR: Surface-Aligned Gaussian Splatting for Efficient 3D Mesh Reconstruction and High-Quality Mesh Rendering},
  author={Gu{\'e}don, Antoine and Lepetit, Vincent},
  journal={CVPR},
  year={2024}
}

@misc{umi,
      title={Universal Manipulation Interface: In-The-Wild Robot Teaching Without In-The-Wild Robots}, 
      author={Cheng Chi and Zhenjia Xu and et.al},
      year={2024},
      eprint={2402.10329},
      archivePrefix={arXiv},
      primaryClass={cs.RO},
      url={https://arxiv.org/abs/2402.10329}, 
}

@InProceedings{dexumi,
  title = 	 {DexUMI: Using Human Hand as the Universal Manipulation Interface for Dexterous Manipulation},
  author =       {Xu, Mengda and Zhang, Han and et.al},
  booktitle = 	 {CoRL},
  pages = 	 {437--459},
  year = 	 {2025},
  editor = 	 {Lim, Joseph and Song, Shuran and Park, Hae-Won},
  volume = 	 {305},
  series = 	 {Proceedings of Machine Learning Research},
  month = 	 {27--30 Sep},
  publisher =    {PMLR},
  url = 	 {https://proceedings.mlr.press/v305/xu25b.html}
}

@article{Articulate_AnyMesh,
    title={Articulate AnyMesh: Open-vocabulary 3D Articulated Objects Modeling}, 
    author={Qiu, Xiaowen and Yang, Jincheng and et.al},
    journal={arXiv preprint arXiv:2502.02590}, 
    year={2025} 
}

@inproceedings{HaMeR,
    title={Reconstructing Hands in 3{D} with Transformers},
    author={Pavlakos, Georgios and Shan, Dandan and et.al},
    booktitle={CVPR},
    year={2024}
}

@misc{pi0.5,
      title={$\pi_{0.5}$: a Vision-Language-Action Model with Open-World Generalization}, 
      author={Physical Intelligence and Kevin Black and et.al},
      year={2025},
      eprint={2504.16054},
      archivePrefix={arXiv},
      primaryClass={cs.LG},
      url={https://arxiv.org/abs/2504.16054}, 
}

@misc{piperx,
  author       = {AgileX Robotics},
  title        = {Piper X: 6-DoF Collaborative Robot Arm},
  year         = {2024},
  howpublished = {\url{https://www.agilex.ai/}},
  note         = {Accessed: 2026-07-22}
}

@inproceedings{lerf,
 author = {Kerr, Justin and Kim, Chung Min and Goldberg, Ken and Kanazawa, Angjoo and Tancik, Matthew},
 title = {LERF: Language Embedded Radiance Fields},
 booktitle = {ICCV},
 year = {2023},
}

@article{groundingdino,
  title={Grounding dino: Marrying dino with grounded pre-training for open-set object detection},
  author={Liu, Shilong and Zeng, Zhaoyang and et.al},
  journal={arXiv preprint arXiv:2303.05499},
  year={2023}
}

@inproceedings{sfm,
    author={Sch\"{o}nberger, Johannes Lutz and Frahm, Jan-Michael},
    title={Structure-from-Motion Revisited},
    booktitle={Conference on Computer Vision and Pattern Recognition (CVPR)},
    year={2016},
}

@inproceedings{mvs,
    author={Sch\"{o}nberger, Johannes Lutz and Zheng, Enliang and et.al},
    title={Pixelwise View Selection for Unstructured Multi-View Stereo},
    booktitle={ECCV},
    year={2016},
}

@misc{postshot,
  title        = {Postshot},
  author       = {{Jawset Visual Computing}},
  year         = {2025},
  howpublished = {\url{https://www.jawset.com/}},
  note         = {Accessed: 2026-05-09}
}

@inproceedings{
lora,
title={Lo{RA}: Low-Rank Adaptation of Large Language Models},
author={Edward J Hu and Yelong Shen and et.al},
booktitle={International Conference on Learning Representations},
year={2022},
url={https://openreview.net/forum?id=nZeVKeeFYf9}
}


\end{document}